\pdfoutput=1
\documentclass[11pt]{article}

\usepackage[]{emnlp2021}

\usepackage{times}
\usepackage{latexsym}
\usepackage[export]{adjustbox}
\usepackage{booktabs}
\usepackage[table,xcdraw]{}
\usepackage[T1]{fontenc}
\usepackage[utf8]{inputenc}

\usepackage{microtype}

\usepackage{multirow}

\title{Privacy Policy Question Answering Assistant \\ A Query-Guided  Extractive  Summarization Approach}  

\author{ Moniba Keymanesh  \\
  The Ohio State University  \\
  \texttt{keymanesh.1@osu.edu} \\ \And
  
 Micha Elsner \\
  The Ohio State University \\
  \texttt{elsner.14@osu.edu} \\ \AND 
  
  Srinivasan Parthasarathy \\
The Ohio State University \\
\texttt{parthasarathy.2@osu.edu}
  
 }

\begin{document}
\maketitle
\begin{abstract}
Existing work on making privacy policies accessible has explored new presentation forms such as color-coding based on the risk factors or summarization to assist users with conscious agreement. 
%However, users often care about a subset of these issues or have a personal view of what is considered risky. Thus, a better alternative is to allow them to ask questions about the issues that they care about and present an answer extracted from the content of the policies.In this work, we take a step toward building an automated privacy policy question-answering assistant. 
To facilitate a more personalized interaction with the policies, in this work, we propose an automated privacy policy question answering assistant that extracts a summary in response to the input user query. This is a challenging task because users articulate their privacy-related questions in a very different language than the legal language of the policy, making it difficult for the system to understand their inquiry. 
Moreover, existing annotated data in this domain are limited.  We address these problems by paraphrasing to bring the style and language of the user's question closer to the language of privacy policies. Our content scoring module uses the existing in-domain data to find relevant information in the policy and incorporates it in a summary. Our pipeline is able to find an answer for 89\%  of the user queries in the privacyQA dataset. 

\end{abstract}

\section{Introduction and Related Work}
\label{intro-query-quided}

%TODO add more related work 
Online users often do not read or understand privacy policies due to the length and complexity of these unilateral contracts~\cite {cranor2006user}. This problem can be addressed by utilizing a presentation form that does not result in cognitive fatigue~\cite{doan2018toward,wurman2001information} and satisfies the information need of users. To assist users with understanding the content of privacy policies and conscious agreement, previous computational work on privacy policies has explored using information extraction and natural language processing to create better presentation forms~\cite{ebrahimi2020mobile}. For example, PrivacyGuide~\cite{tesfay2018privacyguide} and PrivacyCheck~\cite{zaeem2018privacycheck} present an at-a-glance description of a privacy policy by defining a set of privacy topics and assigning a risk level to each topic. Harkous et al~\cite{harkous2018polisis} and Mousavi Nejad et al~\cite{nejad2019towards} used information extraction and text classification to create a structured and color-coded view of the risk factors in the privacy policy. Manor et al~\cite{manor-li-2019-plain} and Keymanesh et al~\cite{keymaneshtoward} explored incorporating the risky data practices in the privacy policies in form of a natural language summary. While great progress has been made to create more user-friendly presentation forms for the policies, users often only care about a subset of these issues or have a personal view of what is considered risky. Instead of presenting an overview or summary of privacy policies, an alternative approach is to allow them to ask questions about the issues that they care about and present an answer extracted from the content of the policies~\cite{ravichanderchallenges}. This facilitates a more personal approach to privacy and enables users to review only the sections of the policy that they are most concerned about.   

In this work, we take a step toward building an automated privacy policy question-answering assistant. We propose to extract an output summary in response to the user query. Our task is related to guided and controllable text summarization~\cite{kryscinski2019neural,dang2008overview, keymanesh2021fairness, fan2017controllable,  sarkhel-etal-2020-interpretable} as well as reading comprehension~\cite{he2020ctrlsum}. However, a few application-imposed constraints make this task more challenging than traditional evaluation setup of reading comprehension systems. First, users tend to pose questions to the privacy policy question-answering system that are not-relevant, out-of-scope~(\textit{‘how many data breaches did you have in the past?’}), subjective~(~e.g. \textit{‘how do I know this app is legit?’}), or too specific to answer using the privacy policy (~e.g. \textit{‘does it have access to financial apps I use?’})~\cite{ravichanderchallenges}. Moreover, even answerable user questions can have a very different style and language in comparison to the legal language used in privacy policies~\cite{ravichander2019question}, making it difficult for the automated assistant to identify the user's intent and find the relevant information in the document. This issue of domain shift is exacerbated due to the difficulty of annotating data for this domain. Because the existing datasets for this task are fairly small~\cite{ahmad2020policyqa}, the problems cannot be solved by simply training a supervised model.   

We address the first problem by using query expansion and paraphrasing to bring the style, language, and specificity of the user's question closer to the language of privacy policies. To do so, we use lexical substitution and back-translation.  Next, using the expanded query-set, we compute a relevance and informativeness score for each segment of the privacy policy using a transformer-based language representation model fine-tuned on in-domain data. Finally, we incorporate the top scored segments in form of a summary.  We show that using a few in-domain datasets annotated for slightly different tasks, we are able extract a relevant summary for 89\% of the user queries in the PrivacyQA dataset.  We discuss our proposed hybrid summarization pipeline in Section~\ref{proposed_pipeline}. We introduce the datasets we use for training and testing different modules of our model in Section~\ref{s:question_quided_summarization_datasets}. Finally, we present our experiments and results in Section~\ref{s:evaluation_and_results}. 
\section{Proposed Query-Guided Extractive Summmarization Pipeline}
\label{proposed_pipeline}

\begin{figure*}[t]
    \centering
  
  \includegraphics[width= 1\linewidth, height = 7
  cm]{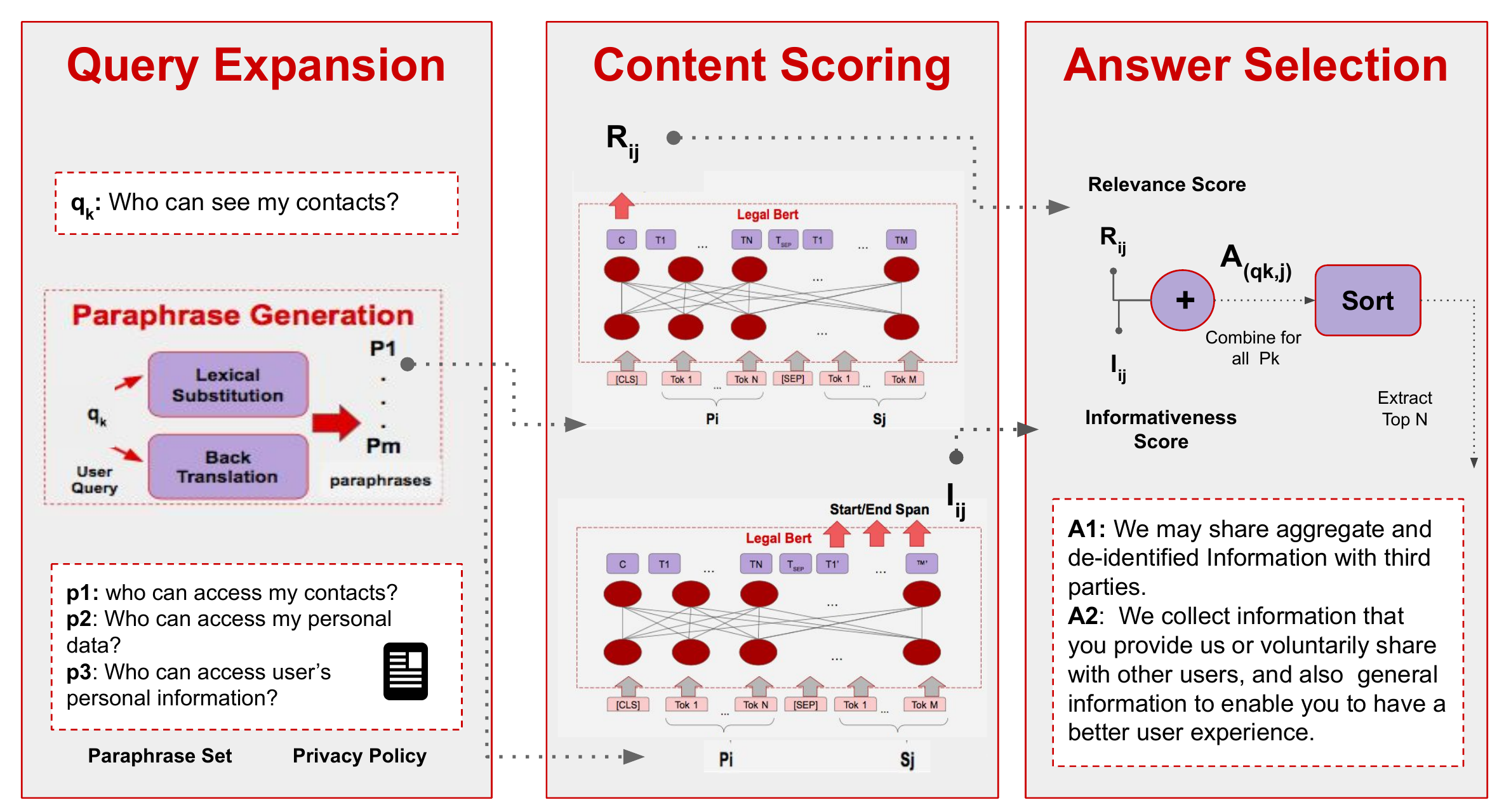} 

\caption{Overview of the proposed pipeline} %\vspace{-0.4cm}
\label{fig:pipeline}
\end{figure*}

In this section, we discuss our proposed pipeline. Our query-guided extractive summarization pipeline includes three main components. Given a privacy policy document and a user query, the first component - the query expansion module, processes the user query and generates a set of paraphrases that have a more similar language, style, and specificity to the content of the privacy policy. Next, given the query and the paraphrase set, the content scoring module computes two scores\textemdash relevance and informativeness\textemdash for each segment of the policy. Lastly, the two scores are combined for the expanded query set to obtain the final answerability score for each segment. Segments are then ranked based on the answerability score and top ranked items are shown in the form of a summary to the user.  The overview of the pipeline as well as an input-output example is shown in Figure~\ref{fig:pipeline}. Next, we discuss the motivation behind including each component in the pipeline and explain them in more detail.

\subsection{Query Expansion}
\label{s:query expansion}
Question-answering systems are very sensitive to many different ways the same information need can be articulated~\cite{dong2017learning}. As a result, small variations in semantically similar queries can yield different answers. This is especially a challenge in building a question-answering assistant for privacy policies. Often, users are not very good at articulating their privacy-related inquiries and use a style and language that is very different from the legal language used in the privacy policies~\cite{ravichanderchallenges}.

Query expansion by paraphrasing has been used in the past to improve the performance of the QA-based information retrieval \cite{zukerman2002lexical, riezler2007statistical, azad2019query}. We employ several methods from the literature, testing their applicability to this domain and in particular to the issues caused by mismatch between external training resources, user queries and the privacy policies themselves. To increase the diversity and coverage of the generated paraphrases, we employ methods based on lexical substitution~\cite{mccarthy2009english, jin2018using} and neural machine translation~(NMT)~\cite{mallinson2017paraphrasing,sutskever2014sequence}. Note that the paraphrase generation module is independent of the neural-based content scoring module and thus, any method can be used to generate paraphrases.  Below, we discuss the three methods used for generating query paraphrases. 

\subsubsection{Lexical Substitution}
Lexical substitution can be done by simply replacing a word with an appropriate synonym/paraphrase in a way that the meaning is not changed. For example, the sentence \textit{'what information is collected about me?'} can be written as \textit{'what information is collected about the user?'}. For generating paraphrases we employ two lexical substitution methods:  1) replacement with similar words based on Word2Vec representations~\cite{le2014distributed}  and 2) a collection of hand-crafted lexical replacements rules aimed to bring the style and language of user queries closer to the legal language in the privacy policies.\footnote{We also tried using WordNet~\cite{miller1998wordnet} for lexical substitution. However, our preliminary experiments suggest that majority of paraphrases generated using this method are not meaningful and thus a more rigorous filtering mechanism should be used to identify useful paraphrases. Thus, we have decided not to use WordNet.}
%\textbf{WordNet: } WordNet is a large lexical database of English. Words are grouped into sets of cognitive synonyms called synsets.  Each synset expresses a distinct concept. Synsets are interlinked by means of conceptual-semantic and lexical relations. We plan to create paraphrases by replacing nouns and verbs with the their synonyms extracted from WordNet. 

\textbf{Word2Vec: } we train the Word2Vec model on a corpus of 150 privacy policies collected by~\citet{keymaneshtoward} to learn word representations. To create paraphrases, we substitute nouns and verbs in user queries with the top 5 most similar words in the embedding space that have the same part of speech. 

\textbf{Lexical replacement rules: } to bring the language and style of user queries closer to the language of privacy policies, we manually create a collection of 50+ lexical substitution rules. For example, our rules can replace the word "my" with "user's" and "phone" with "device". We test two variations of this approach: (i) single-replacement, in which we only apply a single replacement rule to generate a paraphrase and (ii) all-replacement, in which we apply all possible lexical substitution rules to generate a paraphrase.

\subsubsection{NMT-based Paraphrase Generation}
One of the well-known approaches for paraphrase generation is bilingual pivoting~\cite{bannard2005paraphrasing,jia2020ask, jin2018using, dong2017learning}. In this approach, a bilingual parallel corpus is used for learning paraphrases based on techniques from paraphrase-based statistical machine translation~\cite{koehn2003statistical}. Intuitively, two sentences in a source language that translates to the same sentence in a target language can be assumed to have the same meaning. ~\citet{mallinson2017paraphrasing} show how the bilingual pivoting method can be ported into NMT and present a paraphrasing method purely based on neural networks. In our work, we use German as our pivot language following~\citet{mallinson2017paraphrasing}, who suggest that it outperforms other languages in several paraphrasing experiments. We employ a simple back translation method to automatically create paraphrases for user queries using Google Translate~\footnote{https://translate.google.com} which is a mature and publicly available online service to translate user queries from English to German and back from German to English.

%\subsection{Filtering}
%Both lexical substitution and back-translation can lead to paraphrases that are semantically or syntactically incorrect. Using such paraphrases can negatively impact the performance of the downstream question-guided summarization task. Thus, one needs to exclude the incorrect paraphrases before feeding them to the summarization model. We propose to use a language model~\cite{mikolov2010recurrent, bengio2003neural} trained on privacy-policies to get a probability of a paraphrase being valid. We additionally, plan to use a language model based on Transformers~\cite{vaswani2017attention} pre-trained on legal domain~\cite{devlin2018bert, chalkidis2020legal}. We fine-tune this model for the question-answering task on the dataset proposed in~\cite{ravichanderchallenges}. Our preliminary experiments indicate that the model fails to find an answer given a question $q$ and a context $c$ if $q$ has very little word overlap with $c$. We use this fine-tuned model to learn what paraphrases are more likely to yield an answer. This signal will be used to exclude paraphrases that are not useful. 

\vspace{-0.1cm}
\subsection{Content Scoring}
\label{s:content-scoring}

Given the segment set  $S = \{s_1, s_2, ...s_n\}$ of the privacy policy, a user query $q_k$ and paraphrase set  $ P_k = \{p_1, p_2, ..., p_m\}$ obtained by paraphrasing $q_k$ using methods explained in~\ref{s:query expansion}, we aim to  extract the most relevant segments $s_j \in S$ of the privacy policy that fully or partially answer at least one of the paraphrased questions $p_i \in P_k$ and incorporate them in a summary.  To do so, for each pair of paraphrase-segment pair $(p_i ,  s_j)$, we compute two scores that we call the relevancy score  $R_{ij}$ and informativeness score $I_{ij}$ (Both scores are computed using BERT~\cite{devlin2018bert}, but we employ different problem formulations, discussed below). We combine these two scores to get the final answerability score  $A_{ij}$ for paraphrase-segment pair $(p_i, s_j)$. Finally we compute the maximum answerability scores of the paraphrase set $P_k$ to get answerability score of query $q_k$. We represent the answerability score of segment $j$ for query $k$ with $A_{(q_k, j)}$. Next, we discuss how these scores are computed. 

\textbf{Relevance Score: }  To compute the relevance score  $R_{ij}$ for a paraphrase-segment pair $(p_i, s_j)$,  we formulate this as a sentence-pair classification task.   In this task, given a question $p_i$ and segment $s_j \in S$, the goal is to predict whether $s_j$ is relevant to $p_i$. To compute the relevance score, we rely on a transformer-based language representation model~\cite{devlin2018bert} pretrained on legal contracts called \textit{legal-bert~\cite{chalkidis2020legal}~\footnote{the base bert model pretrained on contracts is obtained from https://huggingface.co/nlpaueb/legal-bert-base-uncased}}. We fine-tune this model for sentence pair classification task on the train set of the Privacy QA dataset proposed in ~\cite{ravichanderchallenges} for 3 epochs. PrivacyQA is a corpus of privacy policy segments annotated as \textit{"relevant"} or \textit{"irrelevant"} for a set of user queries. We will further discuss this dataset in Section~\ref{s:question_quided_summarization_datasets}. During fine-tuning, we pass the question-segment pairs separated with the special token $[SEP]$ with question and segment using different segment embeddings. We also add a special token $[CLS]$ in the beginning and a token $[SEP]$ at the end of the sequence. We use weighted binary cross-entropy  as our loss function and update the encoder weights during the fine-tuning. The final hidden vector for the first input token $[CLS]$ is fed to the output layer for the relevance classification task.  We use the fine-tuned model to get the posterior probability of relevancy for each paraphrase-segment sequence.  

\textbf{Informativeness Score: } Even if a segment of the policy is relevant to a question, it might not fully answer it. To account for this,  we also train a span-detection QA system similar to those used for SQUAD question answering~\cite{rajpurkar2016squad}. In preliminary experiments, we do not find that this system always extracts spans which are legible enough on their own for presentation to the user; this is partly due to the complex, contextually-sensitive language used in the contracts. However, we do find that the system's ability to find a promising span provides another indication that the text segment contains a potential answer. Thus, for each question-segment pair $(p_i , s_j)$ we compute an informativeness score which measures how informative is $s_j$ in answering $p_i$. To compute this score, we fine-tune the legal-bert~\cite{chalkidis2020legal} for question-answering task on the train set of the PolicyQA dataset~\cite{ahmad2020policyqa}. This dataset contains reading comprehension style question and answer pairs from a corpus of privacy policies. We will further discuss this dataset in Section~\ref{s:question_quided_summarization_datasets}. We refer to the legal-bert fine-tuned for question-answering as the \textit{"answer-detector" } module in our experiment in Section~\ref{s:results_query_expansion}. During fine-tuning, we feed a query and segment of the policy as a packed sequence separated by the special token $[SEP]$ with the question and the segment using different segment embeddings. In addition, a start vector $S$  and an end vector $E$ are introduced during the fine-tuning process. For each token in the sequence two probabilities are computed: i) the probability of word $k$ being the start of the answer span and ii) the probability of word $k$ being the end of the answer span. To compute the start-of-answer-span probability we compute the dot-product of the token vector $T_k$ and the start vector $S$ followed by a softmax over all tokens in the segment. A similar formula is used to compute the end-of-answer-span probability for each token. The training objective during fine-tuning is to maximize the sum of the log-likelihoods of the correct start and end positions.
The informativenss score of the span from position $a$ to position $b$ is defined as $ S \cdot T_a + E  \cdot T_b $ where $a \leq b$.  We represent the informativeness score of the segment $s_j$ with respect to the paraphrase $p_i$  with $I_{ij}$ and compute it by taking the maximum score of the spans within the segment:
\[
I_{ij} = max( S \cdot  T_a + E \cdot  T_b ) 
\]
Where $0 \leq a \leq b \leq len(s_i)$.

\subsection{Answer Ranking and Selection}
\label{s:answer-selection}

Finally, to compute the answerability score $A_{ij}$ for each paraphrase-segment pair $(p_i, s_j)$, we simply sum up the relevance score $R_{ij}$ and informativeness score $I_{ij}$~\footnote{We also tried training a regression model using $R_{ij}$ and $I_{ij}$ as inputs and the relevance labels from PrivacyQA as the target variable. However, reusing PrivacyQA labels seems to result-in over-fitting. Thus, we decided the combine the scores by simply summing them up.}:
\[
A_{ij} = R_{ij} + I_{ij}
\]
 $A_{ij}$ is computed for all paraphrase-segment pairs $(p_i, s_j)$. However, both lexical substitution and back-translation can generate paraphrases that are semantically or syntactically incorrect. To discard the less useful paraphrases, for each query-segment pair $(q_k, s_j)$ we compute the maximum of the answerability score that paraphrases $ P_k = \{p_1, p_2, ...., p_m\}$ of user query $q_k$ obtained in the previous step~\footnote{We also tried a variation in which we compute the average answerability score of paraphrase set $P_k$. However, our experiments indicated that computing the maximum is more effective as it discards the less useful paraphrases. }:  
\[
A_{(q_k, j)} = max(A_{ij}) ; \forall p_i \in P_k 
\]
Where $P_k$ represents the set of paraphrases generated for user query $q_k$.  Finally, we rank segments $s_j \in S$ based on their answerability score $A_{(q_k, j)}$ with respect to the input user query $q_k$.  The final query-guided summary is built by concatenating the top ranked segments. In our experiments in Section~\ref{s:evaluation_and_results} we show the results of including the top 5 and top 10 ranked segments in the summary. Our query-guided extractive summarization pipeline is shown in Figure~\ref{fig:pipeline}. In the next section, we introduce the datasets used for fine-tuning and testing our proposed pipeline.

\vspace{-0.1cm}
\section{Datasets}
\label{s:question_quided_summarization_datasets}

\begin{table*}[t]\resizebox{1\linewidth}{!}{

\begin{tabular}{@{}lrrrrr@{}}
\toprule
      & \textbf{\#Questions} & \textbf{\#Policies} & \textbf{\begin{tabular}[c]{@{}c@{}}\#out-of-scope \\  questions\end{tabular}} & \textbf{\begin{tabular}[c]{@{}c@{}}Avg. passages\\ per question\end{tabular}} & \textbf{\begin{tabular}[c]{@{}c@{}}Avg. relevant \\ passage per question\end{tabular}} \\ \hline
Train & 1350                 & 27                  & 425                                                                         & 137.1                                                                                        & 5.2                                                                                          \\
Test  & 400                  & 8                   & 34                                                                          & 155.3                                                                                        & 15.5                                                                                        \\ \bottomrule
\end{tabular}}
\caption{Statistics of PrivacyQA Dataset; where \# denotes number of questions, policies, and out-of-scope questions. Out-of-score questions refer to questions for which no segment in the policy is annotated as relevant. We also report the average number of annotated passages/segments and the average number of relevant segments for each question.}
\label{t:prvacyQA_datasets_stats}
\end{table*}

% \usepackage{booktabs}
\begin{table*}[t]\resizebox{1\linewidth}{!}{

\begin{tabular}{@{}lrrrrrr@{}}
\toprule
      & \textbf{Questions} & \textbf{\#Policies} & \textbf{\#Q\&A pairs}  & \textbf{\begin{tabular}[c]{@{}c@{}}Avg. question\\ length\end{tabular}} & \textbf{\begin{tabular}[c]{@{}c@{}}Avg. passage \\ length\end{tabular}} & \textbf{\begin{tabular}[c]{@{}c@{}}Avg. answer\\  length\end{tabular}} \\ \hline
Train &  693              & 75                  &      17,056    & 11.2                                                                    & 106.0                                                                   & 13.3                                                                   \\
Valid &  568           & 20                  &  3,809        & 11.2                                                                    & 96.6                                                                    & 12.8                                                                   \\
Test  &  600               & 20                  &       4,152                            & 11.2                                                                    & 119.1                                                                   & 14.1                                                                   \\ \bottomrule
\end{tabular}}
\caption{Statistics of PolicyQA Dataset;  where \# denotes number of questions, policies, and  Q\&A pairs. We also report the average number of words in passages/segments, questions, and answer spans.}
\label{t:policyQA-dataset-stats}
\end{table*}

We rely on three publicly available data sets for training and testing different modules in our proposed pipeline. As mentioned earlier, we train the word2vec model used for lexical substitution on the set of 150 privacy policies collected by Keymanesh et al~\cite{keymaneshtoward}.  In addition, we employ two datasets called PrivacyQA~\cite{ravichanderchallenges} and PolicyQA~\cite{ahmad2020policyqa} for fine-tuning the legal bert model for sentence pair classification and question-answering tasks respectively. PrivacyQA is a sentence-selection style question-answering dataset where each question is answered with a list of sentences. On the other hand, PolicyQA is a reading-comprehension style question-answering dataset in which a question is answered with a sequence of words. Next, we introduce these datasets in more details.  

\textbf{PrivacyQA:} Ravichander et al~\cite{ravichanderchallenges} asked each crowd worker in their study to formulate 5 privacy questions about privacy policies of a set of 35 mobile applications. The crowd workers were only exposed to the public information about each company. In addition, they were not required to read the privacy policies to formulate their questions. Thus, this dataset presents a more realistic view of what type of questions are likely to be posed to an automotive privacy policy question-answering assistant. Given the questions formulated by Mechanical Turkers, four experts with legal training annotated paragraphs on the privacy policy as \textit{"relevant"} or \textit{"irrelevant"} considering each query. We consider a segment of the privacy policy as relevant if at least one of the annotators marked it as relevant. The datasets statistics is shared in Table~\ref{t:prvacyQA_datasets_stats}. 
We use the train portion of this data set for fine-tuning the legal bert model~\cite{chalkidis2020legal} for the sentence pair classification task and computing the relevance score. We use the test of the PrivacyQA dataset to evaluate our proposed pipeline. We share our results in Section~\ref{s:evaluation_and_results}.

\textbf{PolicyQA:} This dataset is curated by Ahmad et al~\cite{ahmad2020policyqa} and contains 25,017 reading-comprehension style question and answer-span pairs extracted from a corpus of 115 privacy policies~\cite{wilson2016creation}. The train portion of this  dataset contains 693 human-written questions with an average answer length of 13.3 words. To curate this dataset, two domain experts used the triple annotations \{Practice, Attribute, Value\} from the OPP-115 dataset~\cite{wilson2016creation} to come up with the questions. For instance, given the triple annotation \{First Party Collection/Use, Personal Information Type, Contact\} and the corresponding answer span “name, address, telephone number, email address” the annotators formulated questions such as, \textit{"What type of contact information does the company collect?"} and \textit{"Will you use my contact information?"}. Note that during the annotation process, the domain experts were asked to formulate questions given the content of the privacy policy. Therefore, PolicyQA questions are less diverse than PrivacyQA and do not fully reflect a real-world user-interaction with a privacy policy question-answering assistant. Thus, we only use this dataset for fine-tuning the legal bert model~\cite{chalkidis2020legal} for question-answering task and computing the informativeness score. We do not use this dataset for evaluation. The statistics of the PolicyQA dataset are shared in Table~\ref{t:policyQA-dataset-stats}.

\vspace{-0.1cm}
\section{Experiments and Results}
\label{s:evaluation_and_results}
\begin{table*}[t]
\centering
\begin{tabular}{lrrrrr}
\hline

\multicolumn{1}{l}{\textbf{Method}}    & \multicolumn{1}{c}{\textbf{\begin{tabular}[c]{@{}c@{}}Rule-based  (one)\end{tabular}}} & \multicolumn{1}{c}{\textbf{\begin{tabular}[c]{@{}c@{}}Rule-based (all)\end{tabular}}} & \textbf{Back-translation} & \multicolumn{1}{c}{\textbf{Word2Vec}} & \multicolumn{1}{c}{\textbf{All}} \\ \hline
\small \textbf{Average  \#paraphrases}        & 1.4                                                                                    & 0.4                                                                                     & 0.9                      & 2.9                                   & 5.7                              \\

\small \textbf{\%Retrieved relevant segments} & 34.5                                                                                    & 19.8                                                                                    & 25.4                      & 54.0                                  & 67.5                             \\ 

\small \textbf{\%Answerable paraphrases}      & 24.2                                                                                    & 31.3                                                                                    & 25.4                      & 28.9                                  & 27.3                             \\
\hline
\end{tabular}
\caption{The average number of paraphrases that could be generated using each method. The percentage of generated answerable paraphrases for non-answerable queries and the percentage of relevant segments that were answerable using at least one of the generated paraphrases by each method. }
\label{tab:query_expansion_results}
\end{table*}

In this section, we present our experimental results. As stated earlier, given a query and a privacy policy, the first module of our framework- the query expansion module- brings the style and language of the user-queries closer to the language of the privacy policy by paraphrasing. Next, given the paraphrases for the input query, the content scoring and answer selection modules retrieve the most relevant snippets of the privacy policy in form of a summary. In our experiments, we aim to answer the following questions: (i) \textit{Does the query expansion module generate paraphrases that have a closer language than the input query to the privacy policy? } (ii) \textit{If so, what proportion of the generated paraphrases are more answerable than the input user query?} (iii) \textit{Does the proposed pipeline succeed in retrieving the relevant sections of the privacy policy in answer to the user queries?}, and (iv) \textit{Which modules in our pipeline are essential for finding relevant answers to user queries?}

 Our experiments presented in Section~\ref{s:results_query_expansion} answer question one and two. Experiments in Section~\ref{s:results_entire_pipeline} answer question three and four. For our experiments, we rely on the test set of the PrivacyQA dataset as it presents a more realistic user interaction with a privacy policy assistant. This dataset is introduced in Section~\ref{s:question_quided_summarization_datasets}.

\vspace{-0.1cm}
\subsection{Query Expansion Results: } 
\label{s:results_query_expansion}

\begin{table*}[t]
\centering
\begin{tabular}{lrrrrr}\hline

\textbf{Setup}     & \textbf{F@5 } & \textbf{F@10 } & \textbf{P@5}  & \textbf{P@10}  & \textbf{MRR}  \\ \midrule

 \small Full Pipeline & \textbf{80.6}   & \textbf{89.0}      & 39.4          & 32.8           & 0.59       \\

\small $-$ Query Expansion                       & 80.2            & 87.9             & 38.4          & 32.7           & 0.59          \\

 \small $-$ Query Expansion   $-$ Answer-detector                                                                                   & 78.3            & 86.6             & \textbf{41.1} & \textbf{33.8} & \textbf{0.63} \\

\hline 
\end{tabular}
\caption{The performance of different variations of our model in retrieving the relevant segments of the policy in response to user queries. \%F@K represent percentage of queries for which at least one relevant segment was found within the top k ranked items. P@K and MRR represent precision at K and mean reciprocal rank.  }
\label{tab:results-ablation}
\end{table*}

To answer our first and second questions regarding the quality and answerability of the generated paraphrases,  we use lexical substitution methods and back-translation for expanding user queries in the test set of the PrivacyQA dataset. These approaches are introduced in Section~\ref{s:query expansion}. The average number of paraphrases generated by each approach is presented in Table~\ref{tab:query_expansion_results}. On average, using these methods, we can create 5.7 paraphrases for each query. The two variations of the rule-based approach,  the single-replacement and all-replacement generate 1.4 and 0.4 paraphrases on average. Note that for some queries only one substitution rule can be applied and thus, the all-replacement variation does not generate any new paraphrases. The back-translation method creates 0.9 paraphrases on average; using this method may not always generate novel text~\footnote{In this work we only use the NMT architecture used by Google translate and German language. Using more architectures or more target languages can expand the pool of generated paraphrases.}. Word2Vec generates more paraphrases than other methods, generating 2.9 paraphrases on average. 

To measure the language similarity between paraphrase $p_i$ and segment $s_j$, we conduct the following experiment. We hypothesize that the answer-detector model introduced in Section~\ref{s:answer-selection}, can successfully detect the answer span within the relevant segment of the policy if the query has a similar language and style to the privacy policy text. Note that in this problem \textit{Recall} is more important than \textit{Precision}. Meaning that being able to extract all the relevant information from the policy is more crucial than falsely including a few irrelevant sentences in the summary. Our experimental design reflects this domain-imposed requirement.  We pass the paraphrase, segment pair $<p_i, s_j>$ that are annotated as "relevant", to answer-detector model and save the extracted answer span~\footnote{We exclude 34 queries for which there was no relevant information in the policy (out-of-scope questions). }. In cases that the paraphrase $p_i$ and segment $s_j$ do not have a similar language the model typically returns no answer~\footnote{This includes an empty sequence or special token $[CLS]$}.  In our experiments, we observe that for 342 of initial user queries and relevant segment of the policy (around 5.5\% of all pairs),  the answer-detector model can't find the answer span. We interpret this as user queries having a different style and language from the policy text. To answer our first question, we measure the percentage of cases for which the expansion method generated an answerable paraphrase. This is shown in Table~\ref{tab:query_expansion_results} as the \textit{percentage retrieved relevant segments}. As shown in the Table, the rule-based approach (one-replacement) and Word2Vec are able to generate at least one answerable paraphrase for 34.5\% and 54\% of the previously non-answerable cases. Note that these two methods on average generate more paraphrases for each query while back-translation cannot change the language and style of the user query in some cases.  We also observe that 67.5\% of all the non-answerable cases could be answered by at least one of the query expansion approaches. Thus, for better recall,  we include all the expansion methods in our pipeline and filter out the non-answerable paraphrases in the next step. 

To answer our second question regarding the quality of the generated paraphrases, we report the percentage of all generated paraphrases by each method that were answerable in Table~\ref{tab:query_expansion_results}.  Note that different expansion methods generate different number of paraphrases. As presented in the Table, 27.3\% of all the paraphrases were answerable. The all-replacement variation of the rule-based method and Word2Vec generate better quality paraphrases in comparison to back-translation (31.3 and 28.9 in comparison to 25.4). We conjecture that this is due to domain mismatch. The NMT architecture generates high-quality paraphrases, but it is trained on out-of-domain data and therefore has no bias to restate the query in a way that makes it match the privacy policies better. On the other hand, the domain-guided rule-based model and word vectors may be less fluent in paraphrasing, but are trained on in-domain data, which allows them to generate better matches.

\subsection{Query-guided Summarization Results: }
\label{s:results_entire_pipeline}

To answer our third and forth question, we evaluate the performance of  our pipeline in retrieving the relevant segments of the policy given a user query.  Essentially, following the query-expansion, we use the content scoring module to generate the relevance score and informativeness score for each paraphrase-segment pair. This process is discussed in Section~\ref{s:content-scoring}. Finally, the answer ranking module combines these scores for the entire paraphrase set and ranks the segments based on their final answerability score. 

Since we do not have reference summaries for this dataset, we do not use conventional summarization metrics~\cite{lin-2004-rouge, zhang2019bertscore,sellam2020bleurt}, to evaluate our results; rather, we evaluate the ability of the pipeline in retrieving the "relevant" segments of the policy within the top $k$ ranked items.   
We rely on metrics used for the evaluation of information retrieval systems. We report the precision@k (P@K) and the mean reciprocal rank (MRR) of the retrieved relevant segments in Table~\ref{tab:results-ablation}. Precision@k indicates the fraction of the relevant segments in the top $k$ ranked items. We report the average value across queries in the test set. Mean reciprocal rank indicates the multiplicative inverse  of the position of the first relevant segment in the resulted ranking. A perfect ranking system achieves a MRR of 1 by always ranking a relevant segment in the first position.  We also report the percentage of queries in our test set for which at least one relevant snippet was listed at the top k retrieved items (shown as F@k in the Table). We observe that our model is able to retrieve a relevant answer for 80.6\% of the queries within the top 5 results and 89\% of the queries within the top 10 results. Note that 8.5\% of the queries in the test set of PrivacyQA are out-of-scope and cannot be answered solely based on the content of the policy. We also observe that on average, 39.4\% of the top 5 ranked passages are actually relevant. Increasing the summarization budget to include top 10 retrieved passages decreases the precision to 32.8. The MRR of the full pipeline is 0.59 indicating that on average the first relevant item appears in the second position in the ranking or higher. 

To evaluate the contribution of each module in our pipeline we conduct an ablation study. To test the effect of the query-expansion, we use the pipeline without this module for finding answers to user queries. We observe that the percentage of queries answered withing top 5 and top 10 ranked items slightly decreases (-0.4\% and -1.1\% respectively). We conclude that the query expansion module slightly boosts the performance of the model. However, for most queries in the test set the model is already able to find at least one relevant passage without using paraphrases. The advantage of using query-expansion is more noticeable when coverage of all relevant information is more critical and summarization budget is larger.

In our next experiment, in addition to the query-expansion, we also remove the answer-detector from our pipeline. In this version of our model, passages are only ranked based on the relevance score. We notice that this further decreases the F@5 and F@10. However, the precision of the top ranked items and MRR slightly improves. We conclude that both the query expansion and answer-detector components are effective in the ability of our model in finding relevant answers to user queries. However, in cases where short answers are desirable (low summarization budget), content scoring based on only relevancy score would be better. 

\section{Conclusion and Future Work}
In this work, we take a step toward building an automated privacy policy question-answering assistant. This presentation form provides a more personalized interaction with privacy policies in comparison to the previous approaches. We address two main challenges in this domain:  (i) difference between language and style of user queries and the legal language of the privacy polices and (ii) low training resources.  To so do, we propose a query-guided summarization pipeline that first uses lexical substitution and back-translation to bring the language and style of the user queries closer to the language of the policies. Next, we use a language representation model fine-tuned on existing in-domain data to compute a relevancy and informativeness score for each segment in the policy regarding the user query. Finally, the top ranked passages are presented to the user in form of a summary. 

Our proposed pipeline can successfully find the relevant information in the privacy policy for 89\% of the queries in the privacyQA dataset. We observed that using a domain-inspired rule-based approach and training word-vectors on in-domain data is more effective than an out-of-domain NMT-based paraphrase generation approach for bringing the language and style of user queries closer to the language of the privacy policy. However, for a high-recall retrieval system it is better to combine several expansion methods.   In addition, we observed that relying on existing in-domain resources for building a question-answering assistant provides a sufficiently high-recall retrieval system. However, more resources are required for increasing the precision of the ranking system. 

Several issues are left for future work. First, the evaluation of the system using Information Retrieval metrics is insufficient to determine its usefulness in practice. Collection of reference summaries and a user study comparing different methods of interaction with privacy policies could help us determine which methods best meet the needs of real-world users. Second, the proposed method answers user queries based solely on the text of the privacy policy, rendering many user queries unanswerable. 
Using additional legal resources might help to address these out-of-scope questions.

\bibliography{references}
\bibliographystyle{acl_natbib}

%\%section*{Acknowledgments}

%\appendix

%\section{Example Appendix}
%\label{sec:appendix}
%\subsection{Appendices}

%Use \verb|\appendix| before any appendix section to switch the section numbering over to letters. See Appendix~\ref{sec:appendix} for an example.

\end{document}